%% file: main.tex
\documentclass[runningheads]{llncs}

\usepackage{orcidlink}

\usepackage[T1]{fontenc}
\input{math_commands.tex}

\usepackage{graphicx}
\usepackage{float}
\usepackage{subcaption}
\begin{document}
\title{LaFAM: Unsupervised Feature Attribution with Label-free Activation Maps}
%

\author{Aray Karjauv\inst{1,2}\orcidlink{0000-0002-4408-7406}
\and
Sahin Albayrak\inst{1}\orcidlink{0000-0001-5092-4584}
}

\authorrunning{A. Karjauv et al.}
%
\institute{Technische Universität Berlin, 10623 Berlin, Germany
\and GT-ARC gGmbH, 10587 Berlin, Germany
}
\maketitle              

\setcounter{footnote}{0}

\input{content/00_abstract}

\input{content/01_introduction}

\input{content/02_related_work}
\input{content/03_method}

\input{content/04_experiments}
\input{content/05_discussion}

\bibliographystyle{splncs04}
\bibliography{ref}

\newpage

\appendix
\input{content/06_appendix}

\end{document}

%% file: math_commands.tex
\usepackage{amsmath,amsfonts,bm}

\def\eqref#1{equation~\ref{#1}}

\def\1{\bm{1}}

\DeclareMathAlphabet{\mathsfit}{\encodingdefault}{\sfdefault}{m}{sl}
\SetMathAlphabet{\mathsfit}{bold}{\encodingdefault}{\sfdefault}{bx}{n}



%% file: content/00_abstract.tex
\begin{abstract}
    Convolutional Neural Networks (CNNs) are known for their ability to learn hierarchical structures, naturally developing detectors for objects, and semantic concepts within their deeper layers. Activation maps (AMs) reveal these saliency regions, which are crucial for many Explainable AI (XAI) methods. However, the direct exploitation of raw AMs in CNNs for feature attribution remains underexplored in literature. This work revises Class Activation Map (CAM) methods by introducing the Label-free Activation Map (LaFAM), a streamlined approach utilizing raw AMs for feature attribution without reliance on labels. LaFAM presents an efficient alternative to conventional CAM methods, demonstrating particular effectiveness in saliency map generation for self-supervised learning while maintaining applicability in supervised learning scenarios.

\end{abstract}

%% file: content/01_introduction.tex
\section{Introduction}

\begin{figure}[t]
    \centering
    \includegraphics[width=\textwidth]{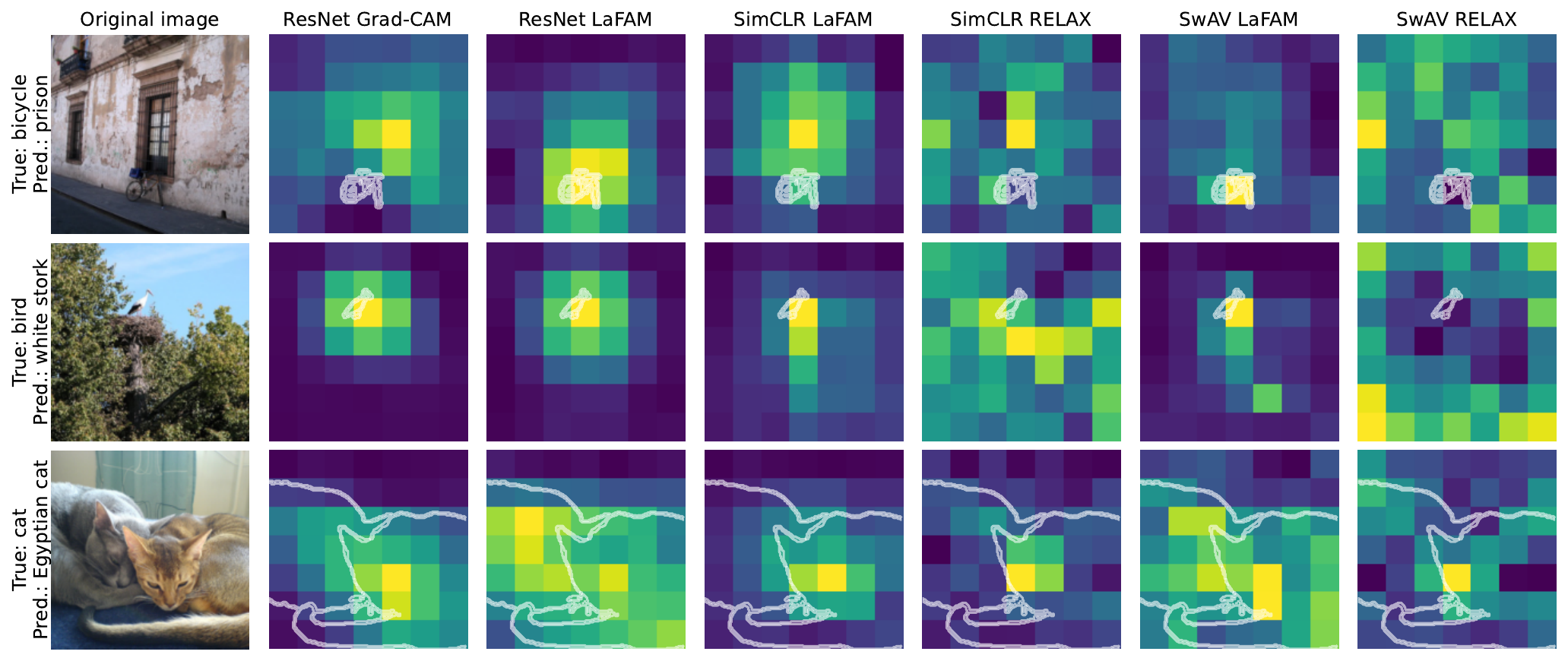}
    \caption{Saliency maps comparison. The contours are overlaid to provide a reference for the ground truth. The title on the left displays the true labels from PASCAL VOC 2012 alongside the ImageNet labels predicted by ResNet50. Grad-CAM utilizes the predicted labels to generate saliency maps. Noticeably, the LaFAM saliency maps are very similar to those produced by Grad-CAM, while RELAX produces noisy saliency maps. The first row demonstrates a misclassification example, showcasing a situation where Grad-CAM fails to highlight the correct region. A detailed comparison is presented in Section \ref{sec:experiment}.}
    \label{fig:results}
\end{figure}

Self-supervised learning (SSL) has emerged as a flexible approach in computer vision~\cite{caron2020unsupervised,grill2020bootstrap}, demonstrating the ability to learn the underlying structure from data without manual annotations. This property enables SSL models to be applied to a variety of downstream tasks, making them often referred to as general purpose or foundation models~\cite{bommasani2021opportunities}. However, the absence of labels presents a challenge in evaluating whether these models have learned relevant representations or if they captured biases and shortcuts~\cite{meehan2023do}. Therefore, it is crucial to identify and address potential shortcomings, particularly in light of the EU AI Act~\cite{aiact}, which requires transparency in high-risk areas such as healthcare.

While Vision Transformers~\cite{dosovitskiy2020image} show impressive results, convolutional neural networks (CNNs) continue to be extensively used due to their well-established effectiveness in extracting spatially coherent features and their interpretability~\cite{smith2023convnets,wightman2021resnet,zeiler2014visualizing}. Evidence suggests that CNNs can inherently develop detectors for semantic objects without explicit supervision~\cite{bau2017network,gonzalez2018semantic,zhou2014object}, making Activation Maps (AMs) in deeper layers a key factor in the success of current XAI methods~\cite{byun2022recipro,chattopadhay2018grad,omeiza2019smooth,ramaswamy2020ablation,selvaraju2017grad,wang2020score,zeiler2014visualizing,zhou2016learning}.

Surprisingly, while AMs are acknowledged for their ability to capture abstract concepts the utilization of raw AMs has been largely overlooked in the literature. Addressing this gap, we propose the Label-free Activation Map (LaFAM) as an alternative to traditional occlusion and gradient-based techniques. LaFAM does not require labels, making it ideally suited for SSL models. It may further benefit supervised learning due to its ability to highlight salient regions that traditional methods might ignore since they primarily focus on generating class-specific saliency maps. For reproducibility and further examination, both the source code\footnote{\url{https://github.com/karray/LaFAM}} and a live demonstration\footnote{\url{https://karay.me/examples/lafam}} are made available.

%% file: content/02_related_work.tex
\section{Related Work}

Early foundational work by Zeiler and Fergus (2014) demonstrates the use of occlusion sensitivity to show how CNN predictions decline when key image features are masked, thereby localizing critical features for classification~\cite{zeiler2014visualizing}. Building on this concept, the Randomized Input Sampling for Explanation (RISE)~\cite{petsiuk2018rise} method employs random masks to create occlusions, facilitating the generation of saliency maps that identify salient regions in image classification tasks.

Zhou et al. (2016) enhanced interpretability by introducing Class Activation Maps (CAMs) which are a linear combination of AMs in the last convolutional layer, multiplied by the weights of the output layer~\cite{zhou2016learning}. Nevertheless, this method is limited by its architecture-specific requirements, which include a global average pooling followed by a fully connected output layer. To mitigate this architectural constraint, various gradient-based~\cite{selvaraju2017grad,chattopadhay2018grad,omeiza2019smooth} and gradient-free methods~\cite{byun2022recipro,ramaswamy2020ablation,wang2020score} were proposed that aim to obtain a class-specific score to weigh the AMs.

However, these methods do not apply to SSL, as they require labels to calculate the class scores. Given the lack of annotations in SSL, there is a requirement for a label-free scoring function that can measure the output. Representation learning explainability (RELAX)~\cite{wickstrom2021relax} method addresses this issue by adapting RISE to SSL models. It utilizes cosine similarity as a replacement for the class score function. It measures the difference between the embedding of the original image and the embeddings of the occluded images, which is then used to weigh the occlusion masks.

Nevertheless, it suffers from several limitations including computational inefficiency due to multiple model inference and noisy results. Additionally, using solid color patches for occlusion may produce out-of-distribution samples, which can produce misleading explanations that do not reflect the behavior of a model on typical in-distribution data.

We propose to utilize the raw AMs directly for generating saliency maps without further manipulations, as they already encapsulate essential semantic and spatial information~\cite{gonzalez2018semantic,zhou2014object}. Given that this approach does not require labeled data, it is particularly well-suited for SSL. Additionally, this method may significantly benefit self-supervised models by revealing all learned concepts, including those not explicitly labeled as shown in Figure~\ref{fig:2_classes}.

%% file: content/03_method.tex
\section{Method}
\label{sec:method}

LaFAM is a post hoc analytical label and gradient-free method that generates saliency maps by averaging the activation maps at a selected convolutional layer. Unlike fully connected layers, convolutional layers preserve spatial information, which is essential for constructing spatially coherent saliency maps. LaFAM requires a single forward pass during inference to obtain the activation of the target convolutional layer, which makes it computationally efficient.

Let $A_{i,j}^{l,k}$ denote the activation at spatial position \((i, j)\) in the $k$-th channel of the $l$-th convolutional layer. The average activation $\bar{A}^l_{i,j}$ over channels at position \((i, j)\) for the $l$-th convolutional layer is computed as follows:

\begin{equation}
    \bar{A}^l_{i,j} = \frac{1}{K}\sum_{k=1}^{K} A_{i,j}^{l,k} \text{,}
\end{equation}
where $K$ is the number of channels in the $l$-th convolutional layer.

The ResNet used in this work incorporates Rectified Linear Units (ReLUs), which ensures that the AMs contain only non-negative values, representing a positive contribution to the output. However, these values are unbounded in the positive direction, requiring normalization. Therefore, the saliency map $M_\text{LaFAM}$ is constructed by first applying min-max normalization to the averaged map $\bar{A}^l$ scaling the values to the range $[0,1]$. The resulting map is then upsampled to match the input image size:

\begin{equation}
    M_\text{LaFAM} = \text{Up} \left( N\left(\bar{A}^l\right) \right) \text{,}
    \label{eq:lafam}
\end{equation}
where $N$ denotes the min-max normalization function, and $\text{Up}$ represents the operation of upsampling the normalized map to the input image size.

%% file: content/04_experiments.tex
\section{Experiment}
\label{sec:experiment}

We systematically evaluate and compare LaFAM with RELAX in SSL settings. To this end, we employ SimCLR~\cite{chen2020simple} and SwAV~\cite{caron2020unsupervised} models with a ResNet50~\cite{he2015deep} backbone, pretrained\footnote{Pretrained models from PyTorch Lightning Bolts} on ImageNet-1k~\cite{russakovsky2015imagenet}. Given that the final convolutional layer produces AMs with a 7x7 spatial dimension and 2048 channels, RELAX is set to generate an equal number of masks with the same spatial resolutions. These masks are produced based on a Bernoulli distribution with a p-parameter of 0.5.

\begin{table}[ht]
    \caption{Saliency maps performance comparison on ImageNet-S (higher values are better).}
    \label{tab:imagenet_results}
    \begin{tabular}{r|c|c||c|c||c|c}
        \hline

                                & \multicolumn{2}{c||}{Supervised (ResNet50)} & \multicolumn{2}{c||}{SSL (SimCLR)} & \multicolumn{2}{c}{SSL (SwAV)}                                           \\
        Metric                  & Grad-CAM                                    & LaFAM                              & RELAX                          & LaFAM          & RELAX & LaFAM          \\
        \hline
        Pointing-Game           & \textbf{94.00}                              & 90.67                              & 88.29                          & \textbf{92.14} & 85.47 & \textbf{89.90} \\
        Sparseness              & \textbf{42.74}                              & 34.82                              & 35.26                          & \textbf{49.70} & 31.49 & \textbf{39.92} \\
        Relevance Mass Accuracy & \textbf{50.28}                              & 45.89                              & 46.19                          & \textbf{53.32} & 42.96 & \textbf{50.13} \\
        Relevance Rank Accuracy & \textbf{62.22}                              & 59.50                              & 58.13                          & \textbf{61.44} & 53.64 & \textbf{64.69} \\
        Top-K Intersection      & \textbf{75.07}                              & 69.09                              & 71.21                          & \textbf{76.59} & 63.83 & \textbf{71.68} \\
        AUC                     & \textbf{83.12}                              & 80.45                              & 76.49                          & \textbf{81.28} & 70.13 & \textbf{83.03} \\
        \hline
    \end{tabular}
\end{table}

This evaluation is conducted using segmentation masks derived from the validation sets of ImageNet-S \cite{gao2022large} and the PASCAL VOC 2012 \cite{everingham2010pascal,everingham2012pascal} datasets. To ensure a fair and consistent comparison, all methods employ nearest neighbor interpolation for upscaling the saliency maps.

Furthermore, we isolate pertinent samples by selecting distinct categories and discarding masks with multiple classes. For the PASCAL VOC 2012 dataset, the image set was narrowed down to 13 labels. The refined datasets consisted of 11,294 and 753 samples from ImageNet-S and PASCAL VOC 2012, respectively.

We also compare the performance of LaFAM and Grad-CAM in supervised settings as a baseline for reference using a pretrained ResNet50 classifier\footnote{Pretrained model from Torchvision}. 

Quantitative analysis is performed using the Quantus framework \cite{hedstrom2023quantus}, with a particular focus on saliency map quantification through segmentation masks. The selected metrics are described in Appendix~\ref{app:quantus}. Visual examples of the saliency maps are presented in Figure~\ref{fig:results}, as well as in Figure~\ref{fig:pascal_more_results} in the Appendix.

\begin{table}[h]
    \caption{Saliency maps performance comparison on PASCAL VOC 2012 (higher values are better).}
    \begin{tabular}{r|c|c||c|c||c|c}
        \hline
                                & \multicolumn{2}{c||}{Supervised} & \multicolumn{2}{c||}{SimCLR} & \multicolumn{2}{c}{SwAV}                                           \\
        Metric                  & Grad-CAM                         & LaFAM                        & RELAX                    & LaFAM          & RELAX & LaFAM          \\
        \hline
        Pointing-Game           & 90.83                            & \textbf{91.23}               & 91.63                    & \textbf{94.68} & 82.73 & \textbf{93.62} \\
        Sparseness              & \textbf{44.39}                   & 36.20                        & 36.04                    & \textbf{51.00} & 32.36 & \textbf{41.61} \\
        Relevance Mass Accuracy & \textbf{40.44}                   & 37.13                        & 38.00                    & \textbf{45.67} & 34.18 & \textbf{42.17} \\
        Relevance Rank Accuracy & 53.53                            & \textbf{53.94}               & 54.55                    & \textbf{58.73} & 46.40 & \textbf{61.05} \\
        Top-K Intersection      & \textbf{65.46}                   & 63.42                        & 67.82                    & \textbf{75.05} & 56.22 & \textbf{67.63} \\
        AUC                     & 82.87                            & \textbf{84.00}               & 79.88                    & \textbf{85.33} & 71.24 & \textbf{87.74} \\
        \hline
    \end{tabular}
    \label{tab:pascal_results}
\end{table}

\begin{figure}[ht]
    \centering
    \includegraphics[width=\textwidth]{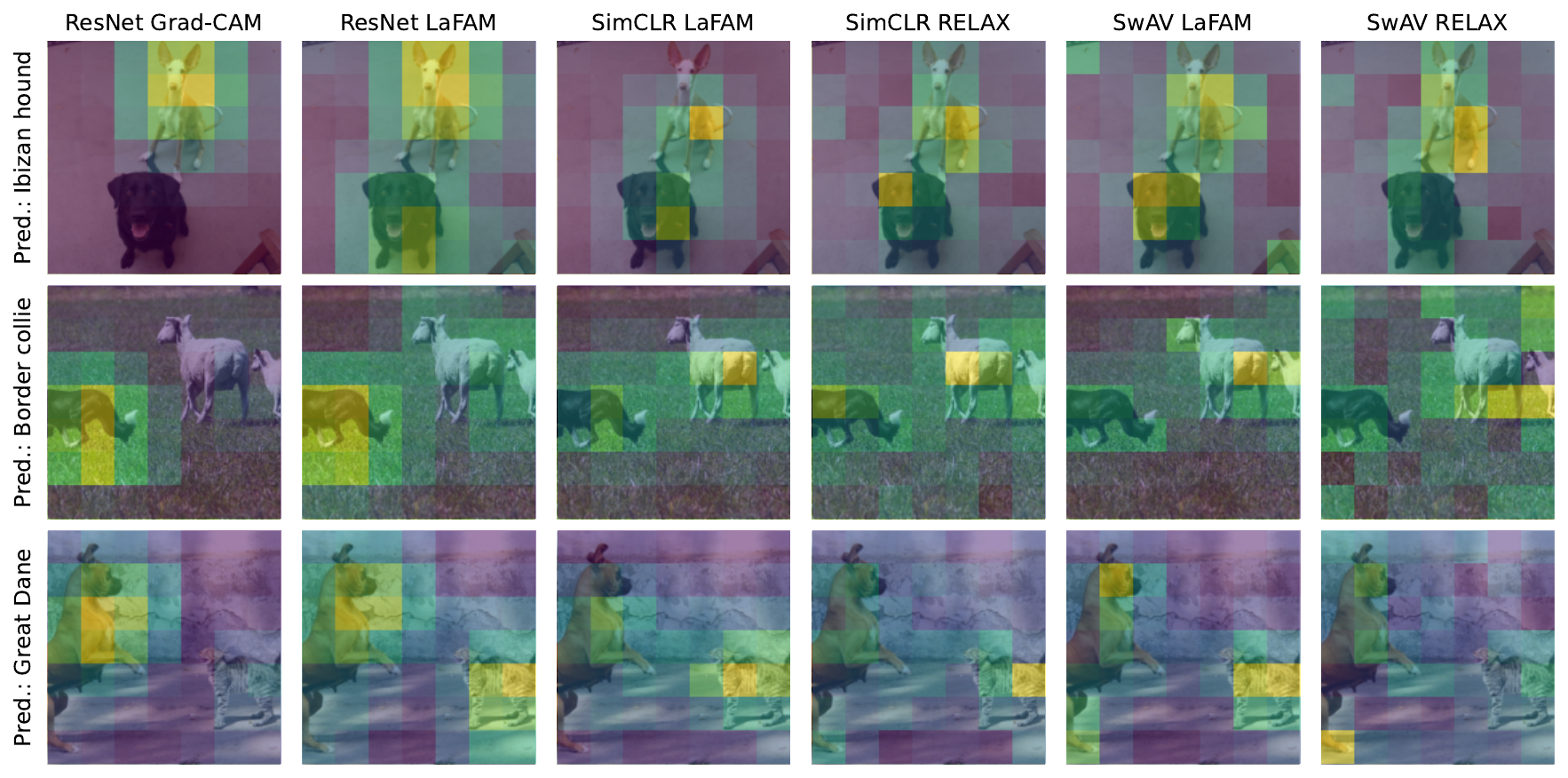}
    \caption{Saliency maps comparison for scenes with two distinct objects. Left-hand labels indicate ImageNet labels predicted by ResNet50 classifier.}
    \label{fig:2_classes}
\end{figure}

Table~\ref{tab:imagenet_results} and Table~\ref{tab:pascal_results} present the results for the ImageNet-1k and PASCAL VOC 2012 datasets, comparing LaFAM with RELAX for SimCLR and SwAV in SSL settings, and with Grad-CAM in the supervised setting as a point of comparison.

In the SSL scenario, LaFAM outperforms RELAX across all metrics for both datasets. Specifically, the higher Pointing-Game and Sparseness scores indicate more precise and focused saliency maps. RELAX shows a low Sparseness score, indicating that it tends to produce noticeably noisy saliency maps, especially for small objects. This observation is supported by the qualitative findings, as depicted in Figure~\ref{fig:results} and Figure~\ref{fig:pascal_more_results} in the Appendix.

In supervised settings, LaFAM is competitive with Grad-CAM across several metrics. As expected, Grad-CAM demonstrates superior results in the Sparseness metric. It implies that Grad-CAM generates less scattered explanations, as its design inherently highlights discriminative regions associated with specific class labels. However, CAM methods require labels and it can be a limiting factor since class scores are computed with respect to the most probable prediction. This can lead to wrong or misleading explanations when the prediction is incorrect.

In contrast, a lower Sparseness score for LaFAM suggests that it captures a broader range of learned features, which can be beneficial compared to CAM methods, especially when an image contains multiple objects in a scene. It, therefore, may not be necessarily interpreted as a shortcoming but rather as a more universal approach as shown in Figure~\ref{fig:2_classes}.

\begin{figure}[ht]
    \centering
    \includegraphics[width=\textwidth]{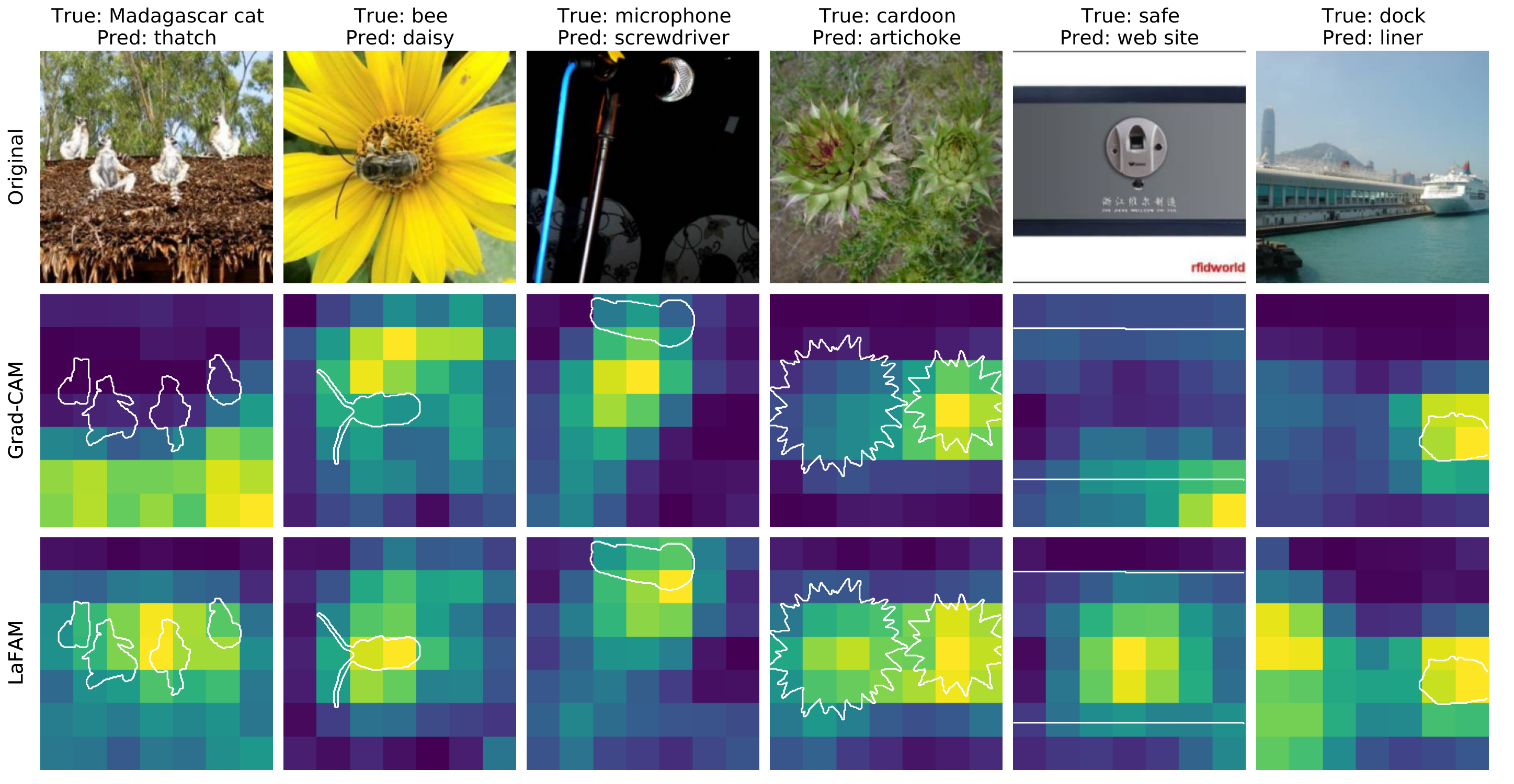}
    \caption{Examples of misclassifications on ImageNet-1k. The left-hand title indicates the method used to generate the saliency maps, while the top title indicates ImageNet-1k ground truth and labels predicted by ResNet50.}
    \label{fig:imagenet_misclassifications}
\end{figure}

Figure~\ref{fig:2_classes} depicts scenes with two distinct objects, demonstrating the ability of LaFAM to highlight multiple concepts simultaneously. This capability is crucial not only for SSL but may also benefit supervised models by providing insights into their behavior when processing complex scenes. Unlike traditional CAM methods, which focus on a single object due to their class-specific approach, LaFAM allows for a more comprehensive understanding of learned concepts. This can be particularly beneficial when predictions are incorrect or when assessing model robustness in real-world situations, where scenes often contain multiple classes. By aggregating AMs without relying on specific predictions, LaFAM can help identify and address model misclassification. 

Figure~\ref{fig:imagenet_misclassifications} shows examples of misclassifications on ImageNet-1k. It can be seen that LaFAM accurately highlights the correct object, in contrast to Grad-CAM. This may indicate that the model has identified the target object, but the presence of multiple objects in the scene may lead to a higher output for an incorrect class. Consequently, the saliency maps produced by Grad-CAM are focused on the wrong object.

%% file: content/05_discussion.tex
\section{Discussion}

This study evaluates methods XAI methods in the context of CNNs and compares the performance of LaFAM and RELAX within the SSL scenario and LaFAM and Grad-CAM under supervised conditions as a reference point.

Our findings reveal that LaFAM consistently outperforms RELAX across all metrics for SSL models, particularly for scenes with small-sized objects. While the performance of RELAX is suboptimal, it can be enhanced by decreasing cell sizes and increasing the number of masks to generate more detailed saliency maps, although this increment raises computational costs. In contrast, LaFAM is more computationally efficient and requires no additional hyperparameters, making it a more practical choice for saliency map generation in SSL scenarios.

A common limitation of traditional CAM methods, including LaFAM, is low resolution. However, it might be possible to address this limitation by redistributing the AM values to earlier layers through Layer-wise Relevance Propagation (LRP)~\cite{bach2015pixel}. This approach could potentially mitigate the issue, but its effectiveness requires further investigation.

In conclusion, the results highlight that LaFAM emerges as a robust and flexible method, contributing to the diversification of the XAI toolbox. Its computational efficiency and label-free saliency map generation make it valuable for understanding model decisions across both supervised and self-supervised learning paradigms.

\section{Acknowledgements}
This work is based on the research conducted in the gemeinwohlorientierter KI-Anwendungen (Go-KI) project (Offenes Innovationslabor KI zur Förderung gemeinwohlorientierter KI-Anwendungen), funded by the German Federal Ministry of Labour and Social Affair (BMAS) under the funding reference number DKI.00.00032.21.

%% file: content/06_appendix.tex
\section{Appendix}

\subsection{Metrics}
\label{app:quantus}

A significant challenge in XAI is the absence of standardized evaluation metrics. Quantus \cite{hedstrom2023quantus} addresses this challenge by offering a flexible toolkit that gathers and systematizes a diverse range of evaluation metrics for explanation methods. Within the framework, we selected the following metrics for evaluation:

\begin{itemize}
    \item \textbf{Pointing-Game} \cite{zhang2018top} establishes whether the most salient attribution aligns with the target object, reflecting the model's precision in highlighting relevant image areas.

    \item \textbf{Top-K Intersection} \cite{theiner2021interpretable} measures the overlap between the binarized explanation and the ground truth mask for the top $k$ features, aiding in assessing alignment with actual object locations.

    \item \textbf{Relevance Rank Accuracy} and \textbf{Relevance Mass Accuracy} \cite{arras2022clevr} evaluate, respectively, the concentration of high-attribution pixels within the ground truth mask and the extent to which positive attributions are confined to the ground truth area.

    \item \textbf{AUC} \cite{fawcett2006introduction} assesses how well attributions correlate with the ground-truth mask by analyzing the receiver operating characteristic curve.

    \item \textbf{Sparseness} \cite{chalasani2020concise} measures explanatory complexity. Using the Gini Index, Sparseness assesses whether the salient features identified are both predictive and free of excess noise, criteria important for interpretability.
\end{itemize}

\begin{figure}[p]
    \centering
    \includegraphics[width=0.9\textwidth]{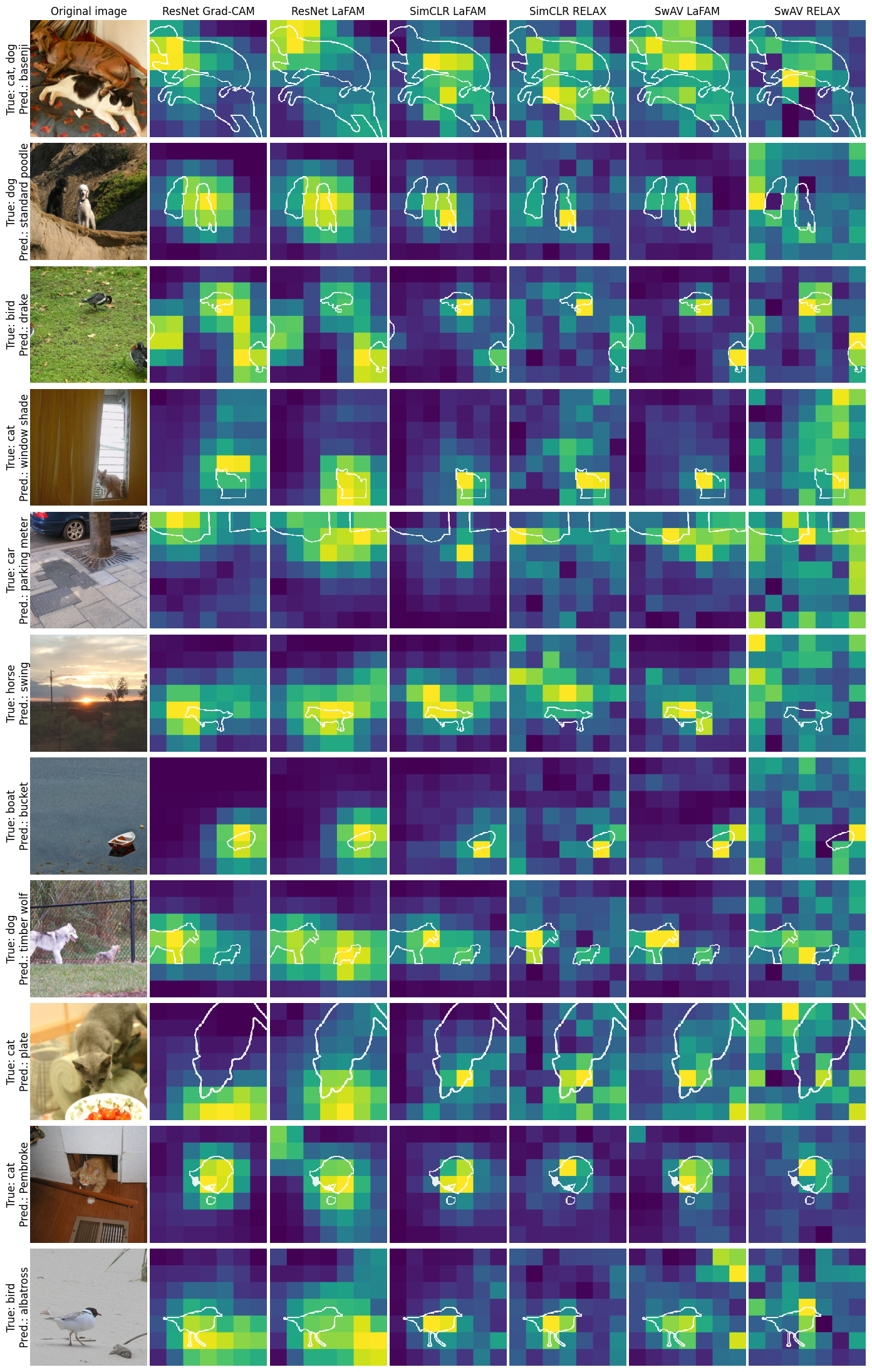}
    \caption{Additional results for PASCAL VOC 2012. The title on the left displays the true labels from PASCAL VOC 2012, alongside ImageNet labels predicted by ResNet50. Grad-CAM utilizes the predicted labels to generate saliency maps.}
    \label{fig:pascal_more_results}
\end{figure}

%% file: main.bbl
\begin{thebibliography}{10}
\providecommand{\url}[1]{\texttt{#1}}
\providecommand{\urlprefix}{URL }
\providecommand{\doi}[1]{https://doi.org/#1}

\bibitem{arras2022clevr}
Arras, L., Osman, A., Samek, W.: {CLEVR-XAI: A benchmark dataset for the ground
  truth evaluation of neural network explanations}. Information Fusion
  \textbf{81},  14--40 (2022)

\bibitem{bach2015pixel}
Bach, S., Binder, A., Montavon, G., Klauschen, F., M{\"u}ller, K.R., Samek, W.:
  On pixel-wise explanations for non-linear classifier decisions by layer-wise
  relevance propagation. PloS one  \textbf{10}(7),  e0130140 (2015)

\bibitem{bau2017network}
Bau, D., Zhou, B., Khosla, A., Oliva, A., Torralba, A.: Network dissection:
  Quantifying interpretability of deep visual representations. In: Proceedings
  of the IEEE conference on computer vision and pattern recognition. pp.
  6541--6549 (2017)

\bibitem{bommasani2021opportunities}
Bommasani, R., Hudson, D.A., Adeli, E., Altman, R., Arora, S., von Arx, S.,
  Bernstein, M.S., Bohg, J., Bosselut, A., Brunskill, E., et~al.: {On the
  opportunities and risks of foundation models}. arXiv preprint
  arXiv:2108.07258  (2021)

\bibitem{byun2022recipro}
Byun, S.Y., Lee, W.: {Recipro-CAM: Gradient-free reciprocal class activation
  map}. arXiv preprint arXiv:2209.14074  (2022)

\bibitem{caron2020unsupervised}
Caron, M., Misra, I., Mairal, J., Goyal, P., Bojanowski, P., Joulin, A.:
  {Unsupervised learning of visual features by contrasting cluster
  assignments}. Advances in neural information processing systems  \textbf{33},
   9912--9924 (2020)

\bibitem{chalasani2020concise}
Chalasani, P., Chen, J., Chowdhury, A.R., Wu, X., Jha, S.: {Concise
  explanations of neural networks using adversarial training}. In:
  International Conference on Machine Learning. pp. 1383--1391. PMLR (2020)

\bibitem{chattopadhay2018grad}
Chattopadhay, A., Sarkar, A., Howlader, P., Balasubramanian, V.N.: {Grad-CAM++:
  Generalized gradient-based visual explanations for deep convolutional
  networks}. In: 2018 IEEE winter conference on applications of computer vision
  (WACV). pp. 839--847. IEEE (2018)

\bibitem{chen2020simple}
Chen, T., Kornblith, S., Norouzi, M., Hinton, G.: {A simple framework for
  contrastive learning of visual representations}. In: International conference
  on machine learning. pp. 1597--1607. PMLR (2020)

\bibitem{dosovitskiy2020image}
Dosovitskiy, A., Beyer, L., Kolesnikov, A., Weissenborn, D., Zhai, X.,
  Unterthiner, T., Dehghani, M., Minderer, M., Heigold, G., Gelly, S., et~al.:
  {An image is worth 16x16 words: Transformers for image recognition at scale}.
  arXiv preprint arXiv:2010.11929  (2020)

\bibitem{everingham2012pascal}
Everingham, M., Van~Gool, L., Williams, C., Winn, J., Zisserman, A.: {The
  PASCAL visual object classes challenge 2012 (VOC2012) results. 2012
  http://www. pascal-network. org/challenges}. In: VOC/voc2012/workshop/index.
  html (2012)

\bibitem{everingham2010pascal}
Everingham, M., Van~Gool, L., Williams, C.K., Winn, J., Zisserman, A.: {The
  PASCAL visual object classes (VOC) challenge}. International journal of
  computer vision  \textbf{88},  303--338 (2010)

\bibitem{fawcett2006introduction}
Fawcett, T.: {An introduction to ROC analysis}. Pattern recognition letters
  \textbf{27}(8),  861--874 (2006)

\bibitem{gao2022large}
Gao, S., Li, Z.Y., Yang, M.H., Cheng, M.M., Han, J., Torr, P.: {Large-scale
  unsupervised semantic segmentation}. IEEE transactions on pattern analysis
  and machine intelligence  (2022)

\bibitem{gonzalez2018semantic}
Gonzalez-Garcia, A., Modolo, D., Ferrari, V.: Do semantic parts emerge in
  convolutional neural networks? International Journal of Computer Vision
  \textbf{126},  476--494 (2018)

\bibitem{grill2020bootstrap}
Grill, J.B., Strub, F., Altch{\'e}, F., Tallec, C., Richemond, P., Buchatskaya,
  E., Doersch, C., Avila~Pires, B., Guo, Z., Gheshlaghi~Azar, M., et~al.:
  {Bootstrap your own latent-a new approach to self-supervised learning}.
  Advances in neural information processing systems  \textbf{33},  21271--21284
  (2020)

\bibitem{he2015deep}
He, K., Zhang, X., Ren, S., Sun, J.: Deep residual learning for image
  recognition. corr abs/1512.03385 (2015) (2015)

\bibitem{hedstrom2023quantus}
Hedstr{\"{o}}m, A., Weber, L., Krakowczyk, D., Bareeva, D., Motzkus, F., Samek,
  W., Lapuschkin, S., H{\"{o}}hne, M.M.M.: {Quantus: An Explainable AI Toolkit
  for Responsible Evaluation of Neural Network Explanations and Beyond}.
  Journal of Machine Learning Research  \textbf{24}(34),  1--11 (2023),
  \url{http://jmlr.org/papers/v24/22-0142.html}

\bibitem{aiact}
Madiega, T.: Artificial intelligence act. European Parliament: European
  Parliamentary Research Service  (2021)

\bibitem{meehan2023do}
Meehan, C., Bordes, F., Vincent, P., Chaudhuri, K., Guo, C.: {Do SSL Models
  Have D\'ej\`a Vu? A Case of Unintended Memorization in Self-supervised
  Learning}. In: Thirty-seventh Conference on Neural Information Processing
  Systems (2023), \url{https://openreview.net/forum?id=lkBygTc0SI}

\bibitem{omeiza2019smooth}
Omeiza, D., Speakman, S., Cintas, C., Weldermariam, K.: {Smooth Grad-CAM++: An
  enhanced inference level visualization technique for deep convolutional
  neural network models}. arXiv preprint arXiv:1908.01224  (2019)

\bibitem{petsiuk2018rise}
Petsiuk, V., Das, A., Saenko, K.: Rise: Randomized input sampling for
  explanation of black-box models. arXiv preprint arXiv:1806.07421  (2018)

\bibitem{ramaswamy2020ablation}
Ramaswamy, H.G., et~al.: {Ablation-CAM: Visual explanations for deep
  convolutional network via gradient-free localization}. In: proceedings of the
  IEEE/CVF winter conference on applications of computer vision. pp. 983--991
  (2020)

\bibitem{russakovsky2015imagenet}
Russakovsky, O., Deng, J., Su, H., Krause, J., Satheesh, S., Ma, S., Huang, Z.,
  Karpathy, A., Khosla, A., Bernstein, M., et~al.: {Imagenet large scale visual
  recognition challenge}. International journal of computer vision
  \textbf{115},  211--252 (2015)

\bibitem{selvaraju2017grad}
Selvaraju, R.R., Cogswell, M., Das, A., Vedantam, R., Parikh, D., Batra, D.:
  {Grad-CAM: Visual explanations from deep networks via gradient-based
  localization}. In: Proceedings of the IEEE international conference on
  computer vision. pp. 618--626 (2017)

\bibitem{smith2023convnets}
Smith, S.L., Brock, A., Berrada, L., De, S.: {ConvNets Match Vision
  Transformers at Scale}. arXiv preprint arXiv:2310.16764  (2023)

\bibitem{theiner2021interpretable}
Theiner, J., M{\"u}ller-Budack, E., Ewerth, R.: {Interpretable semantic photo
  geolocalization}. arXiv preprint arXiv:2104.14995  (2021)

\bibitem{wang2020score}
Wang, H., Wang, Z., Du, M., Yang, F., Zhang, Z., Ding, S., Mardziel, P., Hu,
  X.: {Score-CAM: Score-weighted visual explanations for convolutional neural
  networks}. In: Proceedings of the IEEE/CVF conference on computer vision and
  pattern recognition workshops. pp. 24--25 (2020)

\bibitem{wickstrom2021relax}
Wickstr{\o}m, K.K., Trosten, D.J., L{\o}kse, S., Boubekki, A., Mikalsen,
  K.{\o}., Kampffmeyer, M.C., Jenssen, R.: {RELAX: Representation Learning
  Explainability}. International Journal of Computer Vision  \textbf{131}(6),
  1584--1610 (2023)

\bibitem{wightman2021resnet}
Wightman, R., Touvron, H., J{\'e}gou, H.: {Resnet strikes back: An improved
  training procedure in timm}. arXiv preprint arXiv:2110.00476  (2021)

\bibitem{zeiler2014visualizing}
Zeiler, M.D., Fergus, R.: {Visualizing and understanding convolutional
  networks}. In: Computer Vision--ECCV 2014: 13th European Conference, Zurich,
  Switzerland, September 6-12, 2014, Proceedings, Part I 13. pp. 818--833.
  Springer (2014)

\bibitem{zhang2018top}
Zhang, J., Bargal, S.A., Lin, Z., Brandt, J., Shen, X., Sclaroff, S.: {Top-down
  neural attention by excitation backprop}. International Journal of Computer
  Vision  \textbf{126}(10),  1084--1102 (2018)

\bibitem{zhou2014object}
Zhou, B., Khosla, A., Lapedriza, A., Oliva, A., Torralba, A.: {Object detectors
  emerge in deep scene CNNs}. arXiv preprint arXiv:1412.6856  (2014)

\bibitem{zhou2016learning}
Zhou, B., Khosla, A., Lapedriza, A., Oliva, A., Torralba, A.: {Learning deep
  features for discriminative localization}. In: Proceedings of the IEEE
  conference on computer vision and pattern recognition. pp. 2921--2929 (2016)

\end{thebibliography}
